\documentclass[conference,compsoc]{IEEEtran}
\ifCLASSOPTIONcompsoc
  \usepackage[nocompress]{cite}
\else
  \usepackage{cite}
\fi

\ifCLASSINFOpdf
   \usepackage[pdftex]{graphicx}
   \graphicspath{{images/}}
\else
  \usepackage[dvips]{graphicx}
  \graphicspath{{images/}}
\fi
\usepackage[cmex10]{amsmath}
\usepackage{listings}
\usepackage{color}

\usepackage{microtype}
\usepackage{graphicx}
\usepackage{subcaption}
\usepackage{booktabs} 

\usepackage{fvextra}
\usepackage{algorithmic}
\usepackage{hyperref}
\usepackage{booktabs}
\usepackage{multirow}
\usepackage[table]{xcolor}
\usepackage{hyperref}
\usepackage{amsfonts}
\usepackage{amsmath,amssymb}
\usepackage{natbib}

\definecolor{codegreen}{rgb}{0,0.6,0}
\definecolor{codegray}{rgb}{0.5,0.5,0.5}
\definecolor{codepurple}{rgb}{0.58,0,0.82}
\definecolor{backcolour}{rgb}{0.95,0.95,0.92}
 
\lstdefinestyle{mystyle}{
    backgroundcolor=\color{backcolour},   
    commentstyle=\color{codegreen},
    keywordstyle=\color{magenta},
    numberstyle=\tiny\color{codegray},
    stringstyle=\color{codepurple},
    basicstyle=\footnotesize,
    breakatwhitespace=false,         
    breaklines=true,                 
    captionpos=b,                    
    keepspaces=true,                 
    numbers=left,                    
    numbersep=5pt,                  
    showspaces=false,                
    showstringspaces=false,
    showtabs=false,
    tabsize=2
}

\usepackage{booktabs}
\usepackage{amsmath,amssymb}
\usepackage{subcaption}
\usepackage{xcolor}
\usepackage{multirow}
\usepackage{graphicx}
\usepackage[ruled]{algorithm2e}
\usepackage{placeins}

\SetAlFnt{\small}\SetAlCapFnt{\small}\SetAlCapNameFnt{\small}\SetAlCapHSkip{0pt}

\newcommand{\KPSours}{0.40}                  
\newcommand{\KPSoursNoRefine}{0.89}          

\newcommand{\KPSoursMaskControlProto}{0.51}  

\newcommand{\KPSoursMzero}{29.88}            

\newcommand{\paramsOursTotal}{10.5\,\text{M}} 
\newcommand{\paramsOursMech}{1.5\,\text{M}}  
\newcommand{\paramsCommonEncoder}{9\,\text{M}} 
\newcommand{\paramsBase}{118\,\text{M}}      
\newcommand{\baseFID}{0.081}
\newcommand{\baseTopThree}{0.797}

\newcommand{\ourFID}{0.065}                  
\newcommand{\ourTopThree}{0.799}             
\newcommand{\FIDoursMaskControlProto}{0.064}    
\newcommand{\TopThreeOursMaskControlProto}{0.801}
\newcommand{\skateOursMaskControlProto}{0.0444} 
\newcommand{\trajFiftyOursMaskControlProto}{0.00}
\newcommand{\locFiftyOursMaskControlProto}{0.00}
\newcommand{\FIDoursNoRefine}{0.066}          
\newcommand{\TopThreeOursNoRefine}{0.800}     
\newcommand{\skateOursNoRefine}{0.0447}       
\newcommand{\trajFiftyOursNoRefine}{0.07}     
\newcommand{\locFiftyOursNoRefine}{0.01}      
\newcommand{\FIDoursMzero}{0.076}             
\newcommand{\TopThreeOursMzero}{0.809}        
\newcommand{\skateOursMzero}{0.0425}          
\newcommand{\trajFiftyOursMzero}{27.55}       
\newcommand{\locFiftyOursMzero}{17.05}        
\newcommand{\multiFIDMzero}{0.074}            
\newcommand{\multiTopThreeMzero}{0.812}       
\newcommand{\skateMultiMzero}{0.0433}         
\newcommand{\trajFiftyMultiMzero}{37.04}      
\newcommand{\locFiftyMultiMzero}{22.56}       
\newcommand{\multiFIDMone}{0.054}             
\newcommand{\multiTopThreeMone}{0.806}        
\newcommand{\skateMultiMone}{0.0461}          
\newcommand{\trajFiftyMultiMone}{0.32}        
\newcommand{\locFiftyMultiMone}{0.04}         

\newcommand{\KPSmultiBest}{0.71}             
\newcommand{\multiFID}{0.054}                
\newcommand{\multiTopThree}{0.799}           

\newcommand{\KPSmultiMC}{0.90}               
\newcommand{\multiFIDMC}{0.053}              
\newcommand{\multiTopThreeMC}{0.802}         
\newcommand{\KPSmultiMzero}{40.63}            

\newcommand{\KPSmultiMone}{3.24}              

\newcommand{\skateMultiMC}{0.0486}            
\newcommand{\trajFiftyMultiMC}{0.00}          
\newcommand{\locFiftyMultiMC}{0.00}           

\newcommand{\skateMulti}{0.0483}              
\newcommand{\trajFiftyMulti}{0.00}            
\newcommand{\locFiftyMulti}{0.00}             

\newcommand{\skateOurs}{0.0448}               
\newcommand{\trajFiftyOurs}{0.00}             
\newcommand{\locFiftyOurs}{0.00}              

\newcommand{\CConcatDSFID}{0.059}            
\newcommand{\CConcatDSTopThree}{0.798}       

\newcommand{\KPSpelvis}{0.69}                
\newcommand{\FIDpelvis}{0.064}               
\newcommand{\TopThreepelvis}{0.791}          
\newcommand{\KPShead}{0.37}                  
\newcommand{\FIDhead}{0.060}                 
\newcommand{\TopThreehead}{0.797}            
\newcommand{\KPSlfoot}{0.40}                 
\newcommand{\FIDlfoot}{0.117}                
\newcommand{\TopThreelfoot}{0.790}           
\newcommand{\KPSlwrist}{0.31}                
\newcommand{\FIDlwrist}{0.080}               
\newcommand{\TopThreelwrist}{0.798}          
\newcommand{\KPSrfoot}{0.37}                 
\newcommand{\FIDrfoot}{0.102}                
\newcommand{\TopThreerfoot}{0.794}           
\newcommand{\KPSrwrist}{0.32}                
\newcommand{\FIDrwrist}{0.092}               
\newcommand{\TopThreerwrist}{0.799}          
\newcommand{\KPSavgSingle}{0.41}             
\newcommand{\FIDavgSingle}{0.086}            
\newcommand{\TopThreeAvgSingle}{0.795}       

\newcommand{\KPSkvCConcat}{0.90}             
\newcommand{\FIDkvCConcat}{0.083}            
\newcommand{\TopThreekvCConcat}{0.804}       
\newcommand{\KPSkvCConcatMtwo}{1.01}         
\newcommand{\FIDkvCConcatMtwo}{0.082}        
\newcommand{\TopThreekvCConcatMtwo}{0.801}   

\begin{document}

\title{KV-Control: Parameter-Efficient K/V Injection for Trajectory-Controlled Text-to-Motion}

\author{
\IEEEauthorblockN{
Tengjiao Sun\textsuperscript{1,2,*},
Pengcheng Fang\textsuperscript{1,2,*},
Xiaoyu Zhan\textsuperscript{2, 3}, \\
Yanwen Guo\textsuperscript{3},
Dongjie Fu\textsuperscript{1},
Xiaohao Cai\textsuperscript{1},
Hansung Kim\textsuperscript{2, \textdagger},
}

\IEEEauthorblockA{
\textsuperscript{1}University of Southampton
\textsuperscript{2}Mogo AI Ltd. \quad
\textsuperscript{3}Nanjing University \quad
}


\IEEEauthorblockA{
\textsuperscript{*}Equal contribution. \quad
\textsuperscript{\textdagger}Corresponding author.
}
}

\twocolumn[{%
\renewcommand\twocolumn[1][]{#1}%
\maketitle
\begin{center}
  \includegraphics[width=\textwidth]{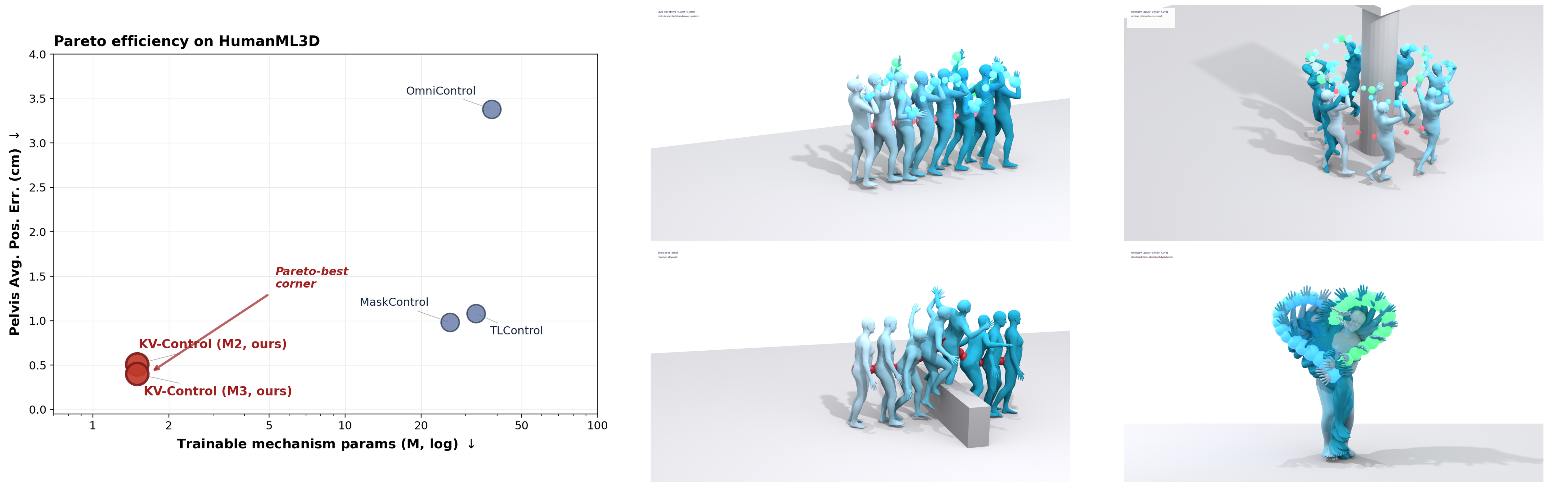}
  \captionof{figure}{\textbf{KV-Control on PartVQ$+$T-Concat.} \emph{Left:} pelvis-trajectory
error versus trainable trajectory-control \emph{mechanism} parameters under the
MaskControl-matched M2 setting; K/V injection uses mechanism
parameters (including the shared trajectory encoder; see
Table~\ref{tab:params}). \emph{Right:} four out-of-distribution demos from the
same trained adapter---\emph{walk\_wave}, \emph{circle\_arms\_high},
\emph{forward\_hop\_wall}, and \emph{walk\_heart\_both\_hands}; markers denote
user-supplied targets, mannequin meshes denote generated motion.
  }
  \label{fig:teaser}
\end{center}
}]

\begin{abstract}
Text-conditioned 3D human motion models now synthesize plausible
motions from prompts, but practical animation and embodied-agent
workflows rarely stop at text: a character may need to follow a sketched
root path, hit an end-effector target, or satisfy a multi-joint trajectory
while still preserving the gait, style, and intent described by language.
This exposes a control trade-off. A trajectory controller should
be precise without overwriting the pretrained text-conditioned motion
prior, yet existing solutions either duplicate large portions of the
generator to regain per-layer control access or move much of the cost to
test-time optimization.
We introduce \textbf{KV-Control}, a compact attention-side control interface for
frozen masked text-to-motion transformers. The key idea is to make
geometric constraints available as memory inside self-attention rather
than injecting them through a global pose token or enforcing them only at
the output side. To support this interface, we co-design a part-tokenized
motion substrate and controller: \textbf{PartVQ} learns anatomy-aligned part
codebooks, \textbf{T-Concat} exposes each frame--part token as an
attention-addressable site, and \textbf{KV-Control} injects control-conditioned
key/value memories at every self-attention layer while preserving the
pretrained query stream, text cross-attention, FFN, and all backbone
weights. The resulting adapter adds only \paramsOursMech{} trainable injection
parameters atop a shared trajectory encoder, yet tracks root and
multi-joint constraints with sub-centimeter accuracy under the inherited
refinement protocol while retaining text-conditioned motion quality.
KV-Control reframes trajectory conditioning as lightweight memory
retrieval, providing a small, precise, and transparent control interface
for text-to-motion generation.
\end{abstract}

\section{Introduction}
\label{sec:introduction}

Text-conditioned 3D human motion generators
\cite{momask2024,mmm2024,motiondiffuse2024,mdm2023,mld2023} now produce diverse, semantically
faithful motion at competitive quality. In production workflows, however, a generated motion is
seldom the end of the pipeline: animators sketch a root path the character must follow,
character-interaction designers fix end-effector contacts to known props, and embodied-AI
planners hand the policy a trajectory the agent must track. A generator may produce a plausible
``walk casually while waving'' when given the prompt alone, yet have the feet drift a dozen
centimetres off the sketched path or the hand miss the prop the user marked---a small text
prompt edit is no help. The unmet need is a generator that is at once \emph{text-faithful} and
\emph{precisely controllable along extrinsic geometric constraints}---and that delivers both
with a control adapter whose trainable parameters and architectural footprint are a small
fraction of the underlying text-to-motion backbone. In this paper we study
\emph{joint trajectory control}: forcing user-specified joint positions---a single
end-effector keyframe, a root path, or a multi-joint trajectory---to match given values
while the prompt continues to dictate gait, style, and intent. Existing
adapters typically reduce to two trained patterns: duplicate-branch adapters
preserve attention access but are heavy, while test-time-optimization
approaches sidestep retraining by adding a per-sample inference-time
refinement stage.

\paragraph{Existing methods and a shared architectural mismatch.}
Two families dominate prior trained trajectory control on text-to-motion
generators, sitting at opposite extremes on a parameters-versus-inference-cost
axis. \textsc{ControlNet}-style adapters \cite{controlnet2023,omnicontrol2024,
maskcontrol2025,intercontrol2024} train a zero-initialized replica of the
backbone connected layer-by-layer to the frozen trunk: this restores
per-layer access to self-attention and keeps text alignment intact, but
the duplicated branch grows the trainable budget to a sizable fraction
of the backbone, scaling roughly with the backbone subset that the branch
duplicates~\cite{omnicontrol2024}. A separate family performs test-time
optimization in latent code or sample noise without a trained
side-adapter---TLControl \cite{tlcontrol2024} refines part-VQ codes via
L-BFGS, DNO \cite{dno2024} optimizes diffusion noise---avoiding the
parameter cost but moving the per-sample compute cost to inference. We
read these two families as instances of an underlying pattern:
\emph{control signal, motion representation, and attention pathway are
architecturally mismatched}---ControlNet restores per-layer attention
access at the cost of duplication, while sample- or code-space optimization
avoids duplication at the cost of per-sample inference compute. Neither
family recovers per-layer attention access at a small fraction of either
cost.

\paragraph{Approach.}
We propose \textbf{KV-Control}, a coupled three-component design in which
a part-tokenized motion representation, a backbone layout, and an
attention-side controller are jointly engineered for parameter-efficient,
per-layer trajectory control on a frozen backbone. The three components
share a single design question: at every
self-attention layer, can each anatomical part-frame token be made
attention-addressable, so a control signal is injected at exactly that
site rather than routed through a global pose token? \textbf{PartVQ}
learns a data-driven $Q$-part anatomical token layout (here
$Q\!=\!6$); \textbf{T-Concat} unpacks these per-part codebooks along the
sequence so each (frame, part) pair is a distinct attention slot with
per-position text cross-attention preserved; the \textbf{K/V injection
module} attaches a control-conditioned key/value pair at each
self-attention layer through a near-identity low-rank projection, leaving
the motion-token query stream and frozen backbone unchanged. The trio is
mutually reinforcing---PartVQ supplies anatomy-aligned tokens, T-Concat
exposes them as attention sites, and the K/V module reads from those
sites with a \paramsOursMech{} mechanism (\paramsOursTotal{} including
the shared \paramsCommonEncoder{} trajectory encoder)---and the coupling
is empirically necessary (\S\ref{sec:exp_base}). PartVQ and T-Concat are
pretrained once and held frozen during the control-adaptation stage that
is the focus of this paper. Figure~\ref{fig:overview} visualizes the
inputs, the trainable adapter, the frozen backbone, and the K/V injection
in a single diagram.

\paragraph{Contributions.}
(1) \textbf{Part-addressable motion substrate}: we introduce a
PartVQ\,$+$\,T-Concat substrate that factorizes motion into
data-driven anatomical part codebooks and unpacks them along the
sequence axis, making each frame--part token an explicit
attention-addressable site while preserving per-position text
cross-attention.
(2) \textbf{Attention-side trajectory control formulation}
(\S\ref{sec:method_kv}): we cast joint trajectory control for frozen
masked motion transformers as memory retrieval inside self-attention,
where continuous joint constraints are encoded as part- and time-addressed
K/V memories rather than injected through global pose tokens or
optimized only at the output side.
(3) \textbf{Parameter-efficient K/V injection mechanism}: we develop
KV-Control, a per-layer near-identity low-rank K/V branch with learnable
control-column attention bias, adding \paramsOursMech{} trainable
injection parameters while leaving the motion-token query stream, FFN,
text cross-attention pathway, and frozen backbone weights unchanged.
Under the MaskControl protocol the system reaches \KPSours\,cm pelvis
error and \KPSmultiBest\,cm multi-joint with \paramsOursMech{} mechanism
parameters --- about $26{\times}$ fewer than a same-backbone
duplicated-branch sanity check at the matched M2 protocol
(Tables~\ref{tab:backbone_ablation}, \ref{tab:trajctrl}).

\section{Related Work}
\label{sec:related}

\paragraph{Text-to-motion generation.}
Masked motion-token models \cite{momask2024,mmm2024,bamm2024,mardm2025}
tokenize motion via VQ-VAE and predict masked tokens, building on the
bidirectional masked-token paradigm introduced for images by MaskGIT
\cite{maskgit2022}; discrete-autoregressive variants \cite{t2mgpt2023}
predict tokens causally instead. Diffusion models
\cite{mdm2023,motiondiffuse2024,mld2023,remodiffuse2023,acmdm2025}
diffuse continuous motion features. Recent work pushes these families along orthogonal axes:
state-space generators \cite{motionmamba2024}, lightweight architectures
\cite{lightt2m2025}, multi-action discrete diffusion \cite{m2d2m2024}, part-coordinating
synthesis \cite{parco2024}, and motion-language LLMs \cite{motiongpt2023}. The two families
reach comparable text-alignment quality; we adopt the masked-transformer family for faster
inference and cheaper adapters. Classical controllable character animation
\cite{motiongraphs2002,pfnn2017,lmm2020} pursued the same runtime-controllability goal in
pre-neural settings.

\paragraph{Motion representation for control.}
A single global VQ-VAE codebook \cite{t2mgpt2023} (or a hierarchical
residual variant \cite{momask2024}) packs whole-body motion into a
monolithic token sequence; TLControl \cite{tlcontrol2024} factors geometry
across manually defined per-part VQs; ACMDM \cite{acmdm2025} argues for
absolute coordinates as inherently control-ready. We follow the part-aware direction but \emph{derive} the partition from data via
lagged cross-correlation clustering with a kinematic-chain integrity prior.

\paragraph{Controllable text-to-motion: a control-interface taxonomy.}
We organize prior trajectory-controlled methods by \emph{where} the control
signal enters the generator. \emph{(i) Sample-, noise-, or code-space
inference-time optimization}: GMD \cite{gmd2023} and PriorMDM
\cite{priormdm2024} steer the iterative sampling trajectory in score- or
sample-space; DNO \cite{dno2024} optimizes diffusion noise latents;
TLControl \cite{tlcontrol2024} refines part-VQ codes via L-BFGS. CondMDI
\cite{condmdi2024} trains a keyframe-conditional in-betweening diffusion
model. MotionLCM \cite{motionlcm2024} accelerates inference orthogonally
to control. \emph{(ii) ControlNet-style duplicated branches}: OmniControl
\cite{omnicontrol2024}, MaskControl/ControlMM \cite{maskcontrol2025}, and
InterControl \cite{intercontrol2024} train a zero-initialized replica of
the backbone connected layer-by-layer to the frozen trunk, restoring
per-layer access at a substantial fraction of the backbone parameter
budget; MaskControl additionally applies logit-side refinement at
inference. \emph{(iii) Semantic body-part editing}: fine-grained
spatio-temporal editing \cite{finemogen2023} and pose-code editing
\cite{como2024} target semantic body-part control rather than spatial
trajectory targets. \emph{(iv) Attention-side memory injection (ours)}:
KV-Control places the trajectory signal as part- and time-addressed
key/value memory \emph{inside} self-attention, between input conditioning
and output logits, leaving the motion-token query stream and frozen
backbone weights unchanged. Inference-time optimization
\cite{maskcontrol2025,tlcontrol2024} is treated as an optional
precision/compute knob (M2/M3 in \S\ref{sec:setup}).

\paragraph{Parameter-efficient adapters.}
KV-Control relates to three families introduced for NLP and image domains:
\emph{adapter modules} \cite{adaptertuning2019} (added bottleneck layers; we instead inject
K/V tokens without inserting new layers), \emph{LoRA} \cite{lora2022} (low-rank weight
updates; we leave all backbone weights untouched), \emph{prefix tuning} \cite{li2021prefix}
(fixed learnable prefixes; ours are dynamically conditioned on per-frame trajectory), and
\emph{IP-Adapter} \cite{ye2023ipadapter} (decoupled cross-attention adding
an image-prompt path to text-to-image diffusion; we fuse into existing
self-attention without inserting additional transformer blocks or
duplicating the trunk).

\section{Method}
\label{sec:method}


\begin{figure*}[!htbp]
    \centering
    \includegraphics[width=0.98\textwidth]{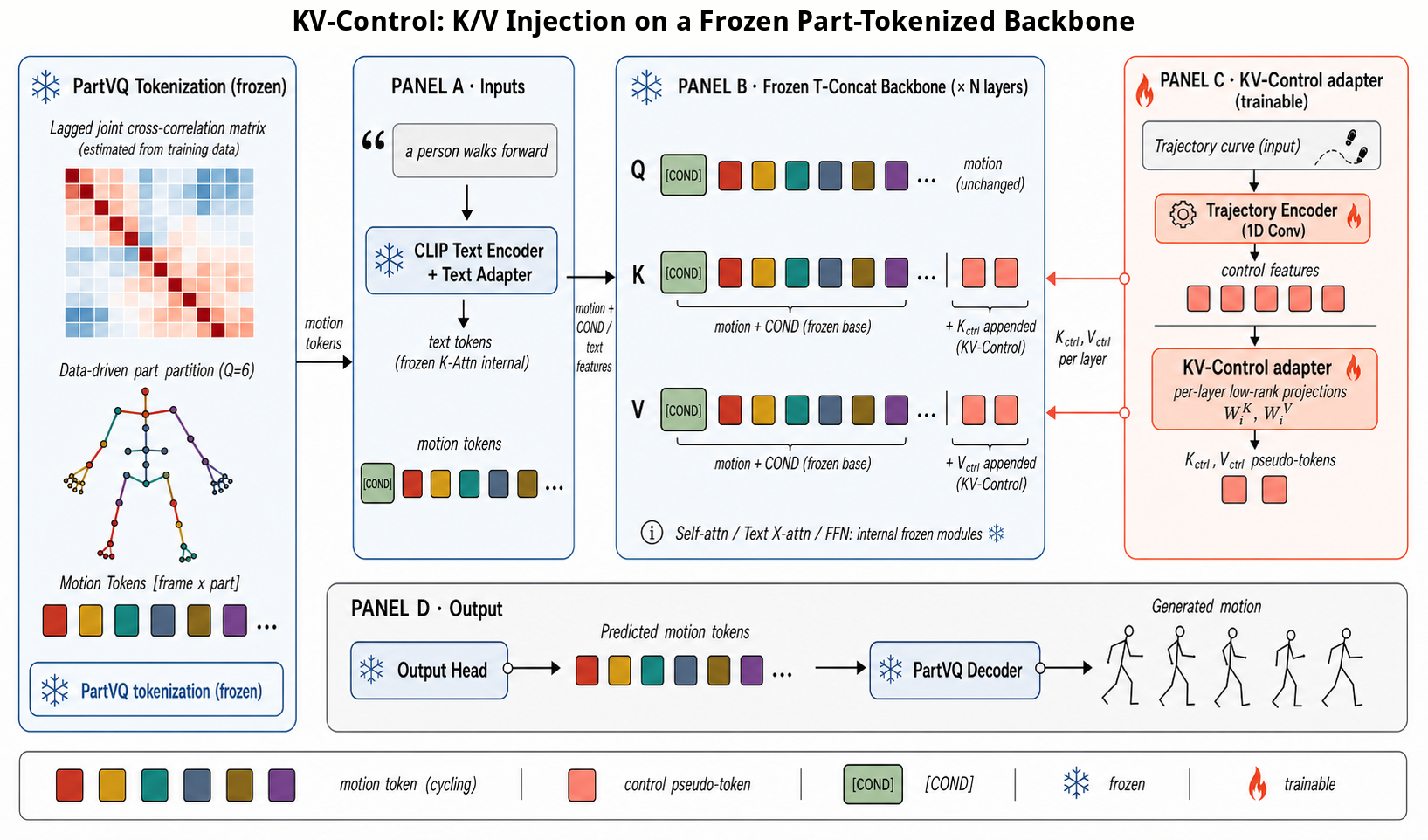}
    \caption{\textbf{KV-Control method overview.} Single-pass left-to-right
    schematic; full equations and dimensions in
    \S\ref{sec:method_partvq}--\S\ref{sec:method_kv}.
    \emph{Left:} frozen co-designed PartVQ$+$T-Concat substrate ($Q\!=\!6$
    data-driven body-part codebooks unpacked along the sequence axis) into
    which motion tokens flow.
    \emph{Middle:} per self-attention layer, the motion query stream $Q$ is
    \emph{unchanged} and the keys/values $K, V$ are augmented with
    control-conditioned pseudo-tokens $K_{\text{ctrl}}, V_{\text{ctrl}}$;
    the Text Cross-Attn and FFN sub-modules also remain frozen.
    \emph{Right:} the trainable control-side module---a shared $1$D-Conv
    trajectory encoder feeds per-layer zero-initialized low-rank
    projections $\mathbf{W}_i^{K},\mathbf{W}_i^{V}$ that produce the
    appended pseudo-tokens.
    \emph{Bottom:} the frozen Output Head and PartVQ Decoder reconstruct
    the final motion.
    \emph{Parameter accounting:} \paramsOursMech{} K/V injection mechanism
    (the control branch compared in Fig.~\ref{fig:teaser});
    \paramsOursTotal{} full control-side trainable budget including the
    \paramsCommonEncoder{} shared trajectory encoder; the underlying
    \paramsBase{} substrate is held strictly frozen during control
    adaptation.}
    \label{fig:overview}%
\end{figure*}

Human motion has structured anatomical degrees of freedom rather than an
undifferentiated token structure. Part trajectories---pelvis, torso, arms,
legs, hands, feet---are coupled but distinct, and external trajectory
targets are typically localized to a small subset of those parts. The
token layout therefore directly affects how easily a control signal can
reach and modulate the relevant degrees of freedom. We exploit this by
exposing part-level structure as sequence positions rather than hiding it
inside the channel dimension, and by routing the control signal into the
attention memory rather than the motion-token query stream.

Concretely, this paper develops three tightly coupled components that
together turn anatomical structure into per-layer attention-addressable
control (Figure~\ref{fig:overview}). \textbf{PartVQ}
(\S\ref{sec:method_partvq}) factors body
geometry across $Q$ per-part codebooks; \textbf{T-Concat}
(\S\ref{sec:method_tconcat}) then unpacks the $Q$ codebooks along the
sequence axis so that each anatomical part-frame token becomes a distinct
attention-addressable site while per-position text cross-attention is
preserved; \textbf{KV-Control} (\S\ref{sec:method_kv}) injects a
control-conditioned key/value pair at each self-attention layer of the
(now-frozen) backbone, reading exactly from the sites the layout exposes.
PartVQ and T-Concat are pretrained once on text-to-motion data and held
frozen during the control-adaptation stage that is the focus of this
paper. \S\ref{sec:method_training} describes the KV-Control training
objective and a matched inference-time refinement adapted from prior
work.

\subsection{Anatomy-Aware Part Tokenization (PartVQ)}
\label{sec:method_partvq}

\paragraph{Partition derivation.}
Per-joint activation $a_j(t)$ is the $\ell_2$ norm of joint $j$'s
relative-to-parent HumanML3D feature, standardized across the corpus.
Pairwise joint similarity is the absolute maximum-lag normalized
cross-correlation $s_{jk}$ over a small lag window; joints are clustered
hierarchically on $1-s_{jk}$ and a kinematic-chain integrity post-step
reassigns each predefined limb chain (left/right arm/leg) to its majority
label so no chain splits.
Algorithm~\ref{alg:partvq} summarizes the full procedure; the
resulting six-part partition used throughout this paper is listed in
the supplement.

\begin{algorithm}[h]
\caption{PartVQ partition derivation (data-driven, anatomy-aligned).}
\label{alg:partvq}
\SetAlgoLined
\SetKwInOut{Input}{Input}\SetKwInOut{Output}{Output}
\Input{Motion corpus $\{M^{(n)}\}_n$ with $M^{(n)}\!\in\!\mathbb{R}^{T_n\times 263}$;\;
parent map $\pi$;\;
limb chains $\mathcal{C}\!=\!\{$l/r-arm, l/r-leg$\}$;\;
lag window $L\!=\!4$;\;
target parts $Q$.}
\Output{Per-joint partition label $\ell\!\in\!\{1,\dots,Q\}^{22}$.}
\BlankLine
\tcc{(1) per-joint activation $a_j(t)$, relative to parent}
\For{$M^{(n)},\, t,\, j$}{
$f_j(t) \leftarrow M^{(n)}_t[\text{ric}_j]-M^{(n)}_t[\text{ric}_{\pi(j)}]$ (raw if $j\!=\!0$)\;
$a_j(t) \leftarrow \|\,f_j(t)\,\|_2$\;
}
$A \leftarrow \mathrm{zscore}(a)$ across all frames\;
\BlankLine
\tcc{(2) lag-aware similarity}
\For{joint pair $(i,j)$}{$s_{ij}\!\leftarrow\!\max_{|\tau|\le L}|\mathrm{corr}(A_i[t],A_j[t{+}\tau])|$}
\BlankLine
\tcc{(3) hierarchical clustering on $1-s$}
$\ell \leftarrow$ \textsc{Linkage}(average) on $\mathrm{condense}(1-s)$, cut at $Q$ clusters\;
\BlankLine
\tcc{(4) kinematic-chain integrity post-step}
\For{$c \in \mathcal{C}$}{$\ell[c]\!\leftarrow\!\mathrm{Majority}(\ell[c])$\;}
\Return $\ell$\;
\end{algorithm}

\paragraph{VQ-VAE architecture and objective.}
A 1D-conv encoder maps each part's motion window to latent tokens quantized by per-part
EMA-reset vector quantizers \cite{vqvae2017}; a shared decoder operates on the
channel-concatenated quantized latents. Training combines smooth-L1 reconstruction, VQ
commitment, and an auxiliary joint-position smooth-L1 loss for direct kinematic supervision.

\subsection{T-Concat Backbone}
\label{sec:method_tconcat}

\paragraph{Motivation.}
A channel-concatenated layout (which we contrast against in
\S\ref{sec:exp_base}) packs the $Q$ part codebooks into a single channel
vector per timestep, leaving downstream attention layers to recover
per-part factors from a shared channel embedding rather than from distinct
sequence positions. T-Concat instead unpacks the codebooks along the
sequence axis, making each anatomical part-codebook factor an
attention-addressable site and giving the K/V branch in
\S\ref{sec:method_kv} per-part landing positions to read from.

\paragraph{Layout.}
T-Concat unpacks the $Q$ codebooks of length $T_{\mathrm{tok}}$ along the sequence axis,
producing $S\,=\,T_{\mathrm{tok}}\!\cdot\!Q$ tokens of fixed dimension $d_{\text{model}}$
(versus a $T_{\mathrm{tok}}$-token channel-concatenated layout used by global-codebook
masked-transformer baselines, which we contrast against in
\S\ref{sec:exp_base}). The $Q$-fold sequence-length increase is the price of
preserved per-quantizer token identity and a one-to-one ratio between motion tokens and
text-conditioning sites.

\paragraph{Position encoding.}
Tokens at position $(t,q)$ in the unpacked sequence receive an additive embedding
$\mathbf{p}_{t,q}\,=\,\mathbf{e}^{(\text{time})}_t\,+\,\mathbf{e}^{(\text{quant})}_q$,
decoupling temporal order from quantizer index without a multiplicative embedding table.

\paragraph{Text pathway and regularization.}
A frozen CLIP \cite{clip2021} text encoder feeds a small TransformerEncoder adapter trained
jointly with the backbone. Its outputs feed gated text cross-attention blocks inserted after
self-attention layers at interval $I$ (we use $I\,=\,1$ in the main configuration so every
self-attention layer is followed by cross-attention). Each cross-attention block has a learnable
scalar gate so its contribution starts near identity. The backbone is a post-norm Transformer
trained with masked-token prediction; we additionally apply a hybrid 2D masking regularizer (a
fraction of samples receive per-quantizer independent masks). Classifier-free-guidance dropout
\cite{ho2022cfg} enables CFG at inference.

\subsection{KV-Control}
\label{sec:method_kv}

\paragraph{Motivation.}
Given the addressable per-part token layout above, the remaining design
question is \emph{how} the trajectory signal should enter the frozen
generator. We supply it as additional attention memory: at each
self-attention layer, control-conditioned pseudo-tokens are appended to
the key/value sets, and the adapter's role reduces to learning what
control memory to add at each layer. The mechanism, the rationale for
choosing the K/V side over the Q side, and the per-layer near-identity
initialization that lets the controller start as a no-op are described
next.

\paragraph{Mechanism.}
KV-Control attaches the control signal as a small set of \emph{additional attention sites} at
the key/value side of every self-attention layer. The motion-token queries and the backbone's
forward pass are unchanged except for a low-rank computation off the side and an additive
attention-bias term. At initialization, the adapter implements a \emph{near-identity} mapping
on the frozen backbone---the control columns participate in the softmax denominator but
contribute negligible mass under the strongly negative initial attention bias---and the
original frozen backbone is recovered \emph{exactly} only in the limit
$b_i\!\to\!-\infty$ (or, equivalently, when the control branch is masked out at inference).
Non-trivial control behavior then emerges as the bias and low-rank weights move away from this
near-identity initialization during training.

\paragraph{Why the key/value side and not the query side.}
The choice is deliberate. A query-side intervention changes the motion-token query stream
that the backbone's text cross-attention reads as per-position grounding, so any disturbance
propagates into how text attends to motion. A key/value-side intervention instead leaves the
motion-token query sequence pointwise unchanged: the control branch contributes additional
entries motion queries may consult, gated to start near zero by the $b_i$ initialization. Concretely, the motion-token query input $\mathbf{Q}_i$
defined below is identical to the frozen backbone's, so the dense per-position text
cross-attention that the backbone was pretrained with is preserved by construction. This
asymmetric treatment of $\mathbf{Q}$ vs $\widetilde{\mathbf{K}},\widetilde{\mathbf{V}}$ makes
K/V injection parameter-efficient and empirically effective under the
PartVQ$+$T-Concat substrate and protocols evaluated here.

\paragraph{Control encoder.}
At each sampling step $s$ the controller decodes the current token state
$\mathbf{z}_s$ via the frozen PartVQ decoder under stop-gradient, producing
$\mathbf{p}^{\text{pred}}_s\,=\,\mathrm{FK}(\mathrm{sg}(D_{\mathrm{VQ}}(\mathrm{safe}(\mathbf{z}_s))))%
\footnote{$\mathrm{safe}(\mathbf{z}_s)$ replaces mask or out-of-codebook ids with a valid zero code before the frozen decoder.}$
(out-of-codebook ids are replaced by $0$). $\mathbf{z}_s$ is the iterative
unmask state at inference and the BERT-style random-masked GT state at
training (the same state consumed by the masked-prediction objective), so
the training residual $\mathbf{p}^{\text{tgt}}-\mathbf{p}^{\text{pred}}_s$
is non-trivial. The control input
$\mathbf{c}_s=[\mathbf{p}^{\text{tgt}}-\mathbf{p}^{\text{pred}}_s;\,\mathbf{p}^{\text{tgt}}]\odot\mathbf{m}$
combines target and residual, gated by the joint--frame mask $\mathbf{m}$.
A 1D-conv stack downsamples to $T_{\mathrm{tok}}$ tokens of dimension
$d_{\text{model}}$. $\mathbf{p}^{\text{pred}}_s$ is read off the same pass
as the transformer, so no extra frozen-base forward is needed
(Table~\ref{tab:protocols} wall-clock includes this).

\paragraph{Per-layer K/V injection.}
Each self-attention layer $i$ is augmented as follows. Let
$\mathbf{X}_i\,\in\,\mathbb{R}^{B\times S\times d_{\text{model}}}$ be the layer's motion-token
input. The shared $\mathbf{f}_{\text{ctrl}}$ is mapped through a layer-specific
\emph{zero-initialized} low-rank down-projection followed by two parallel up-projections,
\[
\mathbf{h}_i\,=\,\mathbf{f}_{\text{ctrl}}\,\mathbf{W}^{\text{down}}_i,\quad
\mathbf{C}^{\text{K}}_i\,=\,\mathbf{h}_i\,\mathbf{W}^{\text{K}}_i,\quad
\mathbf{C}^{\text{V}}_i\,=\,\mathbf{h}_i\,\mathbf{W}^{\text{V}}_i,
\]
with $\mathbf{W}^{\text{down}}_i\,\in\,\mathbb{R}^{d_{\text{model}}\times r}$ initialized to
zero and $\mathbf{W}^{\text{K}}_i,\mathbf{W}^{\text{V}}_i\,\in\,\mathbb{R}^{r\times d_{\text{model}}}$
randomly initialized. Augmented key/value input sequences are formed by concatenation,
$\widetilde{\mathbf{K}}_i^{\text{in}}\,=\,[\,\mathbf{X}_i;\,\mathbf{C}^{\text{K}}_i\,]$ and
$\widetilde{\mathbf{V}}_i^{\text{in}}\,=\,[\,\mathbf{X}_i;\,\mathbf{C}^{\text{V}}_i\,]$
of length $S\,+\,T_{\mathrm{tok}}$. The query input remains $\mathbf{X}_i$, unchanged. The
attention output of layer $i$ is then the standard
\[
\mathrm{Attn}_i\,=\,\mathrm{softmax}\!\Bigl(\tfrac{\mathbf{Q}_i\widetilde{\mathbf{K}}_i^{\!\top}}{\sqrt{d_h}}\,+\,\mathbf{B}_i\Bigr)\,\widetilde{\mathbf{V}}_i,
\]
with $\mathbf{Q}_i\,=\,\mathbf{X}_i\mathbf{W}^{\text{Q}}_i$, $\widetilde{\mathbf{K}}_i\,=\,\widetilde{\mathbf{K}}_i^{\text{in}}\mathbf{W}^{\text{K},\text{attn}}_i$, $\widetilde{\mathbf{V}}_i\,=\,\widetilde{\mathbf{V}}_i^{\text{in}}\mathbf{W}^{\text{V},\text{attn}}_i$ formed from the frozen backbone projections, and $\mathbf{B}_i\,\in\,\mathbb{R}^{S\times(S+T_{\mathrm{tok}})}$ a learnable additive bias that is zero on motion-token columns and a single learnable scalar $b_i$ on the control-token columns, initialized to a strongly negative value so the controller starts near identity at step zero. This double near-identity initialization lets the adapter learn its own layer-wise influence curve from training without a hand-tuned schedule.

\subsection{Frozen-Backbone Training and Inference Refinement}
\label{sec:method_training}

\paragraph{Frozen-backbone discipline.}
During control adaptation, the PartVQ tokenizer, CLIP text adapter, and
T-Concat backbone are pretrained and held fixed; their parameters are
never updated by the control objective. The forward pass still traverses the frozen token
embeddings, transformer self-attention layers, gated text cross-attention, PartVQ decoder, and
the forward-kinematics path that turns codebook embeddings into joint positions, so the
optimizer sees the full computation graph and can route gradients through it. The trainable
budget, however, is limited to the control encoder, the per-layer K/V down/up projections, and
the control-column attention bias scalars. Concretely, the masked-token cross entropy
($\mathcal{L}_{\mathrm{CE}}$) and the trajectory loss ($\mathcal{L}_{\mathrm{traj}}$) below
supervise the controlled output estimate $\hat{\mathbf{x}}^{\text{out}}_0$ produced by the
K/V-augmented transformer; they do \emph{not} supervise the detached current-state estimate
$\mathbf{p}^{\text{pred}}_s$ that the controller consumes when building $\mathbf{c}_s$ in
\S\ref{sec:method_kv}.

\paragraph{Adapter training objective.}
KV-Control is trained with AdamW. The objective combines a masked
cross-entropy term on motion tokens with a forward-kinematics trajectory
loss on the controlled joints,
\[
\mathcal{L}_{\mathrm{KV}}
=
\lambda_{\mathrm{CE}}\mathcal{L}_{\mathrm{CE}}
+
\lambda_{\mathrm{traj}}\mathcal{L}_{\mathrm{traj}},
\]
where $\mathcal{L}_{\mathrm{CE}}$ uses the masked-token loss formulation
of MoMask~\cite{momask2024} on the per-part codebooks, and
\[
\mathcal{L}_{\mathrm{traj}}
=
\|\mathbf{m}\odot(\mathrm{FK}(\hat{\mathbf{x}}^{\text{out}}_0)
-\mathbf{p}^{\text{tgt}})\|_1
\]
applies FK supervision to the predicted clean-token estimates
$\hat{\mathbf{x}}^{\text{out}}_0$. Because the predictions are discrete
codebook tokens, we evaluate $\mathcal{L}_{\mathrm{traj}}$ through a soft
codebook relaxation: token logits define an expectation over each
per-part codebook, which is decoded by the frozen PartVQ decoder before
FK. This differentiable path also supports the optional inference-time
refinement used in M1/M2/M3, where logits or post-sampling token
embeddings are optimized while model parameters remain fixed; the exact
schedules are reported in Table~\ref{tab:protocols}.

\section{Experiments}
\label{sec:experiments}

\subsection{Implementation Details}
\label{sec:setup}
We evaluate on HumanML3D \cite{humanml3d2022} under the MaskControl protocol
\cite{maskcontrol2025} (CFG $3.25$, $T\,=\,10$ steps, seed $3407$, batch $32$). PartVQ uses
$Q\,=\,6$ parts (codebook $128$, code dim $128$). The T-Concat backbone is a $20$-layer
post-norm Transformer ($d_{\text{model}}\,=\,384$, $8$ heads, sequence $S\,=\,294$ tokens)
with frozen CLIP ViT-B/32. KV-Control uses per-layer rank $r\,=\,64$,
$\mathbf{W}^{\text{down}}_i$ zero-init, control bias $b_i\,=\,-5$, contributing
$1{,}474{,}560$ K/V parameters; together with the shared trajectory encoder
the full control-side trainable budget is \paramsOursTotal{}. Inference-time
refinement is adopted verbatim from prior work~\cite{maskcontrol2025,tlcontrol2024}
and exposes two stages---Stage 1 optimizes motion-token logits during sampling, and
Stage 2 optimizes token embeddings post-sampling, both with AdamW
(lr $6{\times}10^{-2}$, $\beta\,=\,(0.5, 0.9)$). Four operating-point configurations
are summarized in Table~\ref{tab:protocols}: \textbf{M0} (pure feed-forward, no
test-time optimization) measures the trained adapter's standalone control ability;
\textbf{M1} adds Stage-1 logit optimization only; \textbf{M2} follows the
MaskControl-matched (MC), iteration-matched refinement schedule; \textbf{M3} is
the strongest-refinement setting and is the operating point used for the
headline numbers. Because M1 and M3 share dynamic Stage-1, M1$\!\rightarrow\!$M3
isolates adding Stage-2; M2 and M3 share Stage-2, so M2$\!\rightarrow\!$M3
isolates the static-to-dynamic Stage-1 switch. Per-stage schedule diagnostics and the
data-driven PartVQ part listing are in the supplemental material.

\begin{table}[!htb]
\caption{Operating points used in evaluation. Stage~1 optimizes
motion-token logits during sampling; Stage~2 refines token embeddings
after sampling. Dynamic Stage~1 uses
$n^{(s)}_{\mathrm{iter}}=(s+1)n_{\mathrm{base}}$ with
$n_{\mathrm{base}}=35$, while static Stage~1 uses a constant per-step
count. Wall-clock is approximate per-sample latency on one H100.}
\label{tab:protocols}
\centering
\footnotesize
\setlength{\tabcolsep}{3pt}
\begin{tabular}{lllcc}
\toprule
Protocol                      & Stage 1                               & Stage 2     & Iters & Wall-clock\,(s) \\
\midrule
\textbf{M0} (feed-forward)    & off                                   & off         & $0$    & $\approx\!1.5$ \\
\textbf{M1} (Stage-1 only)    & dynamic $(s{+}1)\!\cdot\!35$          & off         & $1925$ & $\approx\!12$  \\
\textbf{M2} (MC-matched)      & static $100$ iters/step              & $600$ iters & $1600$ & $\approx\!10$  \\
\textbf{M3} (strongest)       & dynamic $(s{+}1)\!\cdot\!35$          & $600$ iters & $2525$ & $\approx\!15$  \\
\bottomrule
\end{tabular}
\end{table}

\subsection{Backbone-Layout and Mechanism Ablations}
\label{sec:exp_base}
The backbone-layout ablation isolates how PartVQ codebooks are presented to the
transformer. \emph{T-Concat} unpacks $Q\,=\,6$ codebooks along the sequence axis
($S\,=\,294$); \emph{C-Concat} channel-stacks them into one $768$-channel frame
token ($S\,=\,49$). Both backbones share the trunk of \S\ref{sec:setup}, are
trained from scratch under matched protocols, then frozen before attaching an
identical KV-Control adapter.

\paragraph{Findings.}
Table~\ref{tab:backbone_ablation} rules out three alternative readings of the
K/V-side gap. \emph{Backbone quality}: T-Concat and C-Concat match on
unconditional FID and Top-3, so the post-adapter divergence is not
attributable to the trunk. \emph{Refinement protocol}: T-Concat tracks
$\sim\!2.25\!\times$ better at M3 and $\sim\!1.98\!\times$ better at M2,
with Top-3 within $0.01$ of C-Concat across both protocols.
\emph{Parameter budget} (Table~\ref{tab:params}): a same-substrate
duplicated-branch with $26\!\times$ our K/V mechanism budget still
reaches only $1.24\,$cm at M2, versus $\KPSoursMaskControlProto\,$cm
for K/V-Control---added capacity supplies reach, not anatomical
addressability.

\paragraph{Why layout matters.}
T-Concat's advantage over C-Concat (Table~\ref{tab:backbone_ablation}) stems from
a representational property: each (frame, codebook) pair occupies a distinct
attention slot, so a low-rank K/V residual at a query position can selectively
modulate one anatomical sub-token. C-Concat collapses all six codebooks into a
single frame-shared hidden state: the adapter sees $49$ query slots rather than
$294$, and any K/V residual must steer six entangled codebooks through one
latent. This places \emph{representation-layout co-design} at the centre of
building parameter-efficient controllable text-to-motion adapters.

\begin{table}[htbp]
\caption{Backbone-layout and mechanism ablations on the same $20$-layer
trunk. KV-Control is evaluated on T-Concat and C-Concat layouts; the
bottom block is a same-substrate duplicated-branch sanity check at the
MaskControl-matched M2 protocol. Bold indicates the main configuration.}
\label{tab:backbone_ablation}
\centering
\footnotesize
\setlength{\tabcolsep}{3pt}
\begin{tabular}{llccc}
\toprule
Backbone                       & Cfg.        & FID                                                & Top-3                                                          & Avg.\,Pos.\,Err.\,(cm)                                    \\
\midrule
\multicolumn{5}{l}{\emph{Unconditional generation (no control adapter)}}                                                                                                                                                                                                \\
T-Concat                       & Base        & $\baseFID$                                         & $\baseTopThree$                                                & --                                                        \\
C-Concat                       & Base        & $\CConcatDSFID$                                    & $\CConcatDSTopThree$                                           & --                                                        \\
\midrule
\multicolumn{5}{l}{\emph{KV-Control adapter (identical configuration), root-trajectory control}}                                                                                                                                                                        \\
T-Concat                       & M2          & $0.064$                                            & $0.801$                                                        & $\KPSoursMaskControlProto$                                \\
C-Concat                       & M2          & $\FIDkvCConcatMtwo$                                & $\TopThreekvCConcatMtwo$                                       & $\KPSkvCConcatMtwo$                                       \\
\textbf{T-Concat (main)}       & \textbf{M3} & $\mathbf{\ourFID}$                                 & $\mathbf{\ourTopThree}$                                        & $\mathbf{\KPSours}$                                       \\
C-Concat                       & M3          & $\FIDkvCConcat$                                    & $\TopThreekvCConcat$                                           & $\KPSkvCConcat$                                           \\
\midrule
\multicolumn{5}{l}{\emph{ControlNet branch on T-Concat ($39.3\,$M / $48.3\,$M full)}}                                                                                                                                                                                       \\
T-Concat                       & M2          & $0.277$                                            & $0.767$                                                        & $1.24$                                                    \\
\bottomrule
\end{tabular}
\end{table}

\begin{table}[!htb]
\caption{Parameter accounting on the same frozen $118\,$M T-Concat
masked-transformer motion backbone. Mechanism column $=$ trajectory-control
mechanism (Fig.~\ref{fig:teaser} x-axis); Full $=$ Mechanism $+$ shared
trajectory encoder.}
\label{tab:params}
\centering
\footnotesize
\setlength{\tabcolsep}{6pt}
\begin{tabular}{lccc}
\toprule
Adapter on T-Concat           & Mechanism\,(M) & Encoder\,(M)   & Full\,(M)         \\
\midrule
\textbf{KV-Control (ours)}    & $\mathbf{1.5}$ & $9.0$          & $\mathbf{10.5}$   \\
ControlNet (dup.\ branch)     & $39.3$         & $9.0$          & $48.3$            \\
\bottomrule
\end{tabular}
\end{table}

\subsection{Trajectory Control}
\label{sec:exp_traj}

The same K/V-Control adapter applies to two trajectory-control settings on
HumanML3D: \emph{single-joint} (pelvis only) and \emph{multi-joint} (random
subset of up to six joints from $\{$pelvis, head, $\ell$/$r$ foot,
$\ell$/$r$ wrist$\}$, sampled per training example). Table~\ref{tab:trajctrl} reports prior-work rows verbatim from
\cite{maskcontrol2025} as benchmark context, and our M1/M2/M3 rows as
re-runs on the KV-Control substrate under the operating points in
Table~\ref{tab:protocols}; same-backbone mechanism evidence is
isolated in Table~\ref{tab:backbone_ablation}.

\paragraph{Findings.}
On single-joint pelvis, M3 reaches $\KPSours\,$cm (FID $\ourFID$,
Top-3 $\ourTopThree$); M1 ($\KPSoursNoRefine\,$cm) uses dynamic Stage-1
without Stage-2; M0 ($\KPSoursMzero\,$cm) is pure feed-forward, vs MaskControl's
$40.41\,$cm at its matched no-optimization
configuration~\cite[Table~5]{maskcontrol2025}. The all-joints
multi-joint M3 row reaches $\KPSmultiBest\,$cm at $17\!\times\!$ fewer
mechanism parameters; the gap to MaskControl's $0.72\,$cm prior
baseline is within the benchmark's reporting precision, so we
interpret it as comparable tracking accuracy rather than a
significant improvement.

\begin{table*}[!t]
\caption{Trajectory control on HumanML3D under the MaskControl evaluation protocol.
Upper/lower blocks: single-joint pelvis / all-joints multi-joint.
\emph{Avg.\,Pos.\,Err.\,(cm)} = mean Euclidean distance between generated and
target positions over controlled joint-frame targets (shared across all rows);
Skate = Foot-Skating Ratio; Traj.\,E, Loc.\,E = $>\!50\,$cm failure rates (\%);
MC = MaskControl-matched compute.}
\label{tab:trajctrl}
\centering
\scriptsize
\setlength{\tabcolsep}{3pt}
\begin{tabular}{lcccccc}
\toprule
Method                              & Top-3 $\uparrow$         & FID $\downarrow$    & Skate $\downarrow$         & Traj.\,E $\downarrow$       & Loc.\,E $\downarrow$       & Avg.\,Pos.\,Err.\,(cm) $\downarrow$            \\
\midrule
\multicolumn{7}{l}{\emph{Train on Pelvis (single-joint), prior-work \cite{maskcontrol2025}}}                                                                                                                                                                                  \\
GMD \cite{gmd2023}                  & $0.665$                  & $0.576$             & $0.1009$                   & $9.31$                      & $3.21$                     & $14.39$                            \\
OmniControl \cite{omnicontrol2024}  & $0.687$                  & $0.218$             & $0.0547$                   & $3.87$                      & $0.96$                     & $3.38$                             \\
MotionLCM \cite{motionlcm2024}      & $0.752$                  & $0.531$             & --                         & $18.87$                     & $7.69$                     & $18.97$                            \\
TLControl \cite{tlcontrol2024}      & $0.779$                  & $0.271$             & --                         & $0.00$                      & $0.00$                     & $1.08$                             \\
MaskControl \cite{maskcontrol2025}  & $\mathbf{0.809}$         & $\mathbf{0.061}$    & $0.0547$                   & $0.00$                      & $0.00$                     & $0.98$                             \\
MaskControl (w/o Logits Opt.) \cite{maskcontrol2025} & $0.802$        & $0.128$             & $0.0594$                   & $39.14$                     & $24.00$                    & $40.41$                            \\
\midrule
\multicolumn{7}{l}{\emph{Train on Pelvis, our re-runs}}                                                                                                                                                                                                                       \\
Ours, M0 (feed-forward)             & $\TopThreeOursMzero$     & $\FIDoursMzero$     & $\skateOursMzero$          & $\trajFiftyOursMzero$       & $\locFiftyOursMzero$       & $\KPSoursMzero$                    \\
\midrule
Ours, M1 (Stage-1)                  & $\TopThreeOursNoRefine$  & $\FIDoursNoRefine$  & $\skateOursNoRefine$       & $\trajFiftyOursNoRefine$    & $\locFiftyOursNoRefine$    & $\KPSoursNoRefine$                 \\
\textbf{Ours, M2 (MC,\,$+$Stage-2)} & $\TopThreeOursMaskControlProto$ & $\FIDoursMaskControlProto$ & $\skateOursMaskControlProto$ & $\trajFiftyOursMaskControlProto$ & $\locFiftyOursMaskControlProto$ & $\mathbf{\KPSoursMaskControlProto}$ \\
\textbf{Ours, M3 (dyn\,Stage-1\,$+$\,Stage-2)} & $\ourTopThree$           & $\ourFID$           & $\skateOurs$               & $\trajFiftyOurs$            & $\locFiftyOurs$            & $\mathbf{\KPSours}$                \\
\midrule\midrule
\multicolumn{7}{l}{\emph{Train on All Joints (multi-joint), prior-work \cite{maskcontrol2025}}}                                                                                                                                                                               \\
OmniControl \cite{omnicontrol2024}  & $0.693$                  & $0.310$             & $0.0608$                   & $6.17$                      & $1.07$                     & $4.04$                             \\
TLControl \cite{tlcontrol2024}      & $0.782$                  & $0.256$             & --                         & $0.00$                      & $0.00$                     & $1.11$                             \\
MaskControl \cite{maskcontrol2025}  & $\mathbf{0.805}$         & $0.083$             & $0.0545$                   & $0.00$                      & $0.00$                     & $0.72$                             \\
\midrule
\multicolumn{7}{l}{\emph{Train on All Joints, our re-runs (multi-joint K/V-Control)}}                                                                                                                                                                                         \\
Ours, M0 (feed-forward)             & $\multiTopThreeMzero$    & $\multiFIDMzero$    & $\skateMultiMzero$         & $\trajFiftyMultiMzero$      & $\locFiftyMultiMzero$      & $\KPSmultiMzero$                   \\
\midrule
Ours, M1 (Stage-1)                  & $\multiTopThreeMone$     & $\multiFIDMone$     & $\skateMultiMone$          & $\trajFiftyMultiMone$       & $\locFiftyMultiMone$       & $\KPSmultiMone$                    \\
Ours, M2 (MC,\,$+$Stage-2)          & $\multiTopThreeMC$       & $\multiFIDMC$       & $\skateMultiMC$            & $\trajFiftyMultiMC$         & $\locFiftyMultiMC$         & $\KPSmultiMC$                      \\
\textbf{Ours, M3 (dyn\,Stage-1\,$+$\,Stage-2)} & $\multiTopThree$         & $\mathbf{\multiFID}$ & $\mathbf{\skateMulti}$    & $\trajFiftyMulti$           & $\locFiftyMulti$           & $\mathbf{\KPSmultiBest}$           \\
\bottomrule
\end{tabular}
\end{table*}

\paragraph{Per-joint decomposition.}
Table~\ref{tab:perjoint} reports the multi-joint adapter with the controlled set
fixed to one anatomical joint at a time. Aggregate single-joint position error is
$\KPSavgSingle\,$cm (FID $\FIDavgSingle$, Top-3 $\TopThreeAvgSingle$);
per-joint position error ranges from $\KPSlwrist\,$cm (left wrist) to $\KPSpelvis\,$cm
(pelvis). The foot rows (left foot FID $\FIDlfoot$, right foot FID $\FIDrfoot$)
breach the $0.10$ FID gate---a reminder that the average single-joint FID should
not be read as uniform per-joint gate satisfaction.

\paragraph{Extended diagnostics.}
Table~\ref{tab:diag_full} adds matching-score (Match., lower better) and
intra-set Diversity (Div., real-data $\sim\!9.5$) to the operating-point
progression, plus a refinement-only baseline (frozen T-Concat, no K/V
adapter). Stripping the K/V adapter and running the identical M2/M3
schedules drives position error to near zero but collapses FID to
$\sim\!104$ and Top-3 to $\sim\!0.12$---the adapter, not the refinement
loop, supplies the structural conditioning that keeps refinement inside
the text-conditioned motion manifold.

\begin{table}[!htb]
\caption{Extended diagnostics under the MaskControl protocol; means
unless noted. Match. is the motion-text matching score of
\cite{humanml3d2022}; Div. is intra-set feature variance with real-data
Diversity~$\sim\!9.5$.}
\label{tab:diag_full}
\centering
\footnotesize
\setlength{\tabcolsep}{3pt}
\begin{tabular}{lccccc}
\toprule
Operating point                & Avg.\,Pos.\,Err.\,(cm)         & FID                  & Top-3                    & Match.\,$\downarrow$ & Div.\,$\rightarrow$ \\
\midrule
\multicolumn{6}{l}{\emph{Single-joint pelvis checkpoint selected by validation Avg.\,Pos.\,Err.}}                                                                                                                \\
M0 (no refinement)             & $\KPSoursMzero$                & $\FIDoursMzero$      & $\TopThreeOursMzero$     & $2.890$              & $9.566$             \\
M1 (Stage-1 only)              & $\KPSoursNoRefine$             & $\FIDoursNoRefine$   & $\TopThreeOursNoRefine$  & $2.945$              & $9.477$             \\
M2 (MC-matched)       & $\KPSoursMaskControlProto$     & $\FIDoursMaskControlProto$ & $\TopThreeOursMaskControlProto$ & $2.939$ & $9.481$             \\
\textbf{M3 (+ refinement)}     & $\mathbf{\KPSours}$            & $\mathbf{\ourFID}$   & $\mathbf{\ourTopThree}$  & $\mathbf{2.946}$     & $\mathbf{9.444}$    \\
\midrule
\multicolumn{6}{l}{\emph{Multi-joint checkpoint selected by validation Top-3}}                                                                                                                                  \\
M0 (no refinement)             & $\KPSmultiMzero$               & $\multiFIDMzero$     & $\multiTopThreeMzero$    & $2.866$              & $9.566$             \\
M1 (Stage-1 only)              & $\KPSmultiMone$                & $\multiFIDMone$      & $\multiTopThreeMone$     & $2.840$              & $9.502$             \\
M2 (MC-matched)       & $\KPSmultiMC$                  & $\multiFIDMC$        & $\multiTopThreeMC$       & $2.934$              & $9.465$             \\
\textbf{M3 (+ refinement)}     & $\mathbf{\KPSmultiBest}$       & $\mathbf{\multiFID}$ & $\mathbf{\multiTopThree}$& $\mathbf{2.939}$     & $\mathbf{9.415}$    \\
\midrule
\multicolumn{6}{l}{\emph{Refinement-only baseline: frozen T-Concat, no K/V adapter}}                                                                                                                            \\
Frozen T-Concat (M2)           & $0.013$                        & $103.86$             & $0.122$                  & $10.049$             & $1.377$             \\
Frozen T-Concat (M3)           & $0.012$                        & $104.40$             & $0.122$                  & $10.059$             & $1.343$             \\
\bottomrule
\end{tabular}
\end{table}

\begin{table}[!htb]
\caption{Per-joint single-joint control on the multi-joint adapter (M3 protocol,
controlled set fixed to one anatomical joint at inference). Each row holds the
controlled set fixed to one anatomical joint; the Average row is the unweighted
mean of the six rows above; the all-joints multi-joint M3 row is repeated from
Table~\ref{tab:trajctrl} for direct comparison.}
\label{tab:perjoint}
\centering
\footnotesize
\setlength{\tabcolsep}{3pt}
\begin{tabular}{lccc}
\toprule
Controlled joint              & FID                  & Top-3                  & Pos.\,Err.\,(cm) \\
\midrule
Pelvis                        & $\FIDpelvis$         & $\TopThreepelvis$      & $\KPSpelvis$     \\
Head                          & $\FIDhead$           & $\TopThreehead$        & $\KPShead$       \\
Left foot                     & $\FIDlfoot$          & $\TopThreelfoot$       & $\KPSlfoot$      \\
Right foot                    & $\FIDrfoot$          & $\TopThreerfoot$       & $\KPSrfoot$      \\
Left wrist                    & $\FIDlwrist$         & $\TopThreelwrist$      & $\KPSlwrist$     \\
Right wrist                   & $\FIDrwrist$         & $\TopThreerwrist$      & $\KPSrwrist$     \\
\midrule
Average (6 single-joint)      & $\FIDavgSingle$      & $\TopThreeAvgSingle$   & $\KPSavgSingle$  \\
\midrule
Multi-joint (all joints, M3)  & $\multiFID$          & $\multiTopThree$       & $\KPSmultiBest$  \\
\bottomrule
\end{tabular}
\end{table}

\subsection{Qualitative Results}
\label{sec:exp_qual}

\begin{figure*}[!p]
\centering
\includegraphics[width=0.96\textwidth]{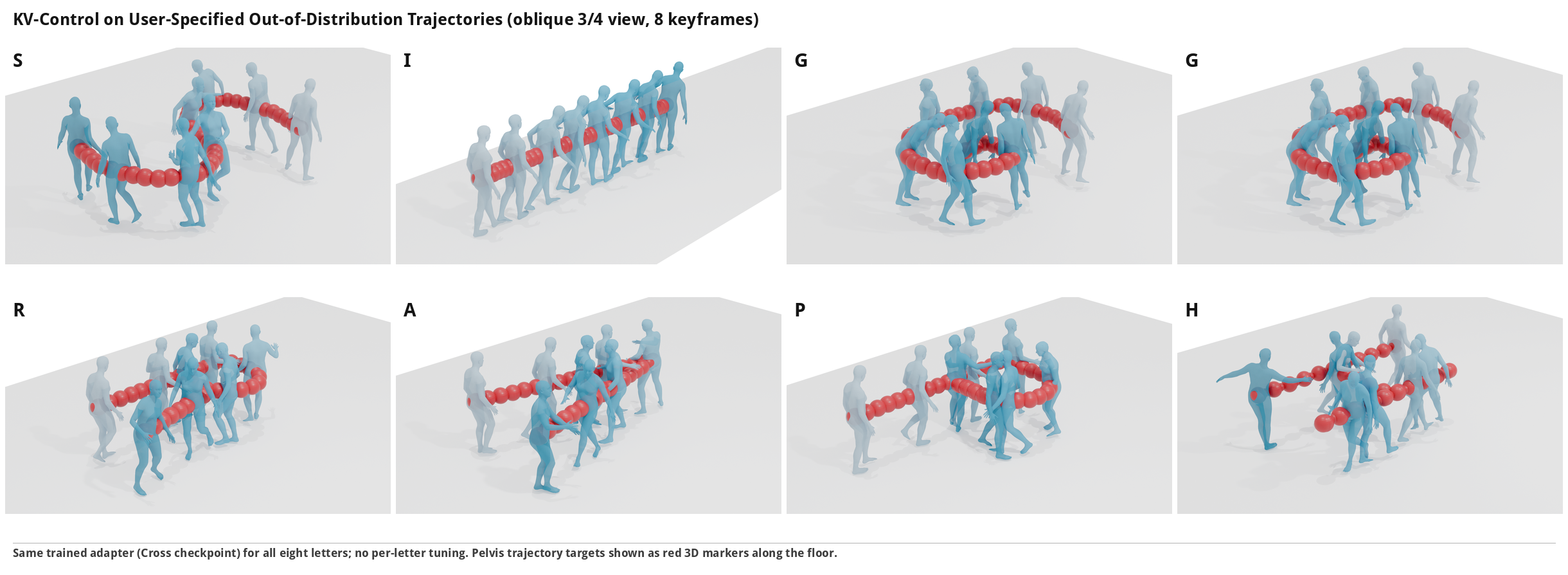}
\caption{\textbf{Qualitative trajectory control on out-of-distribution targets.}
Eight user-specified letter-shaped pelvis-trajectory targets (S, I, G, G, R, A, P, H)
for our single-joint pelvis K/V-Control checkpoint, oblique $3/4$ view; each cell
overlays input waypoints as red 3D markers on the floor along $8$ translucent body
keyframes sampled from the $L\,=\,196$ motion. The same trained adapter is applied to
all eight letters with no per-letter tuning; the duplicate G panel is for visual
readability of the SIGGRAPH word. $20$\,fps per-letter videos are in the supplementary.}
\label{fig:qual}
\end{figure*}

\begin{figure*}[!p]
\centering
\includegraphics[width=0.97\textwidth]{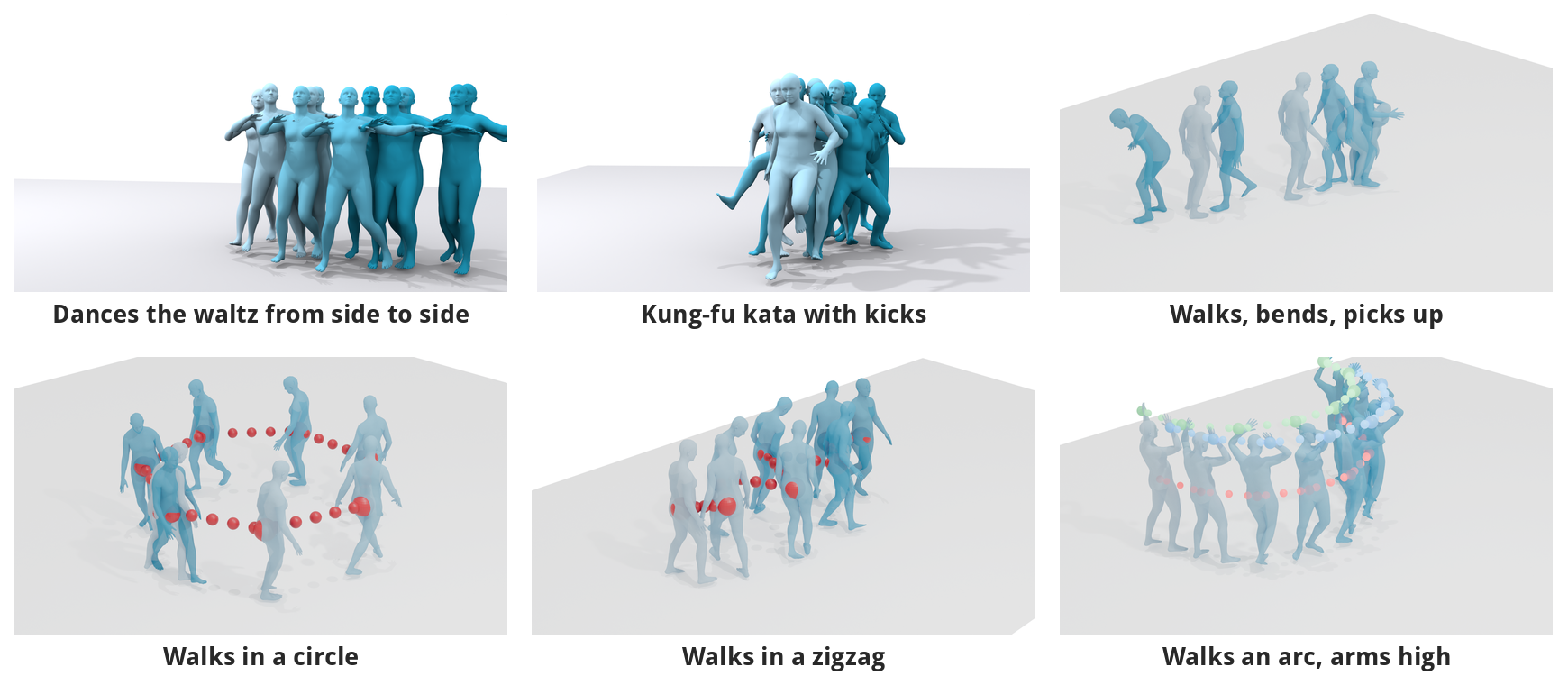}
\caption{\textbf{Unconditional generation (top) vs.\ trajectory-controlled
synthesis (bottom).} Top: the frozen base backbone on dynamic text prompts
(waltz side-step, kung-fu kata with kicks, walk/bend/pick-up) synthesizes
coherent multi-stage motion, confirming that Table~\ref{tab:trajctrl} gains
sit atop a competent generator. Bottom: the K/V-Control checkpoint applied
to three out-of-distribution constraint patterns---walks in a circle, walks
in a zigzag, walks an arc with arms held high (multi-joint). Colored 3D
markers are the supplied control targets (red\,$=$\,pelvis,
blue\,$=$\,l\_wrist, green\,$=$\,r\_wrist); per-demo videos in the
supplementary.}
\label{fig:demos}
\end{figure*}

Figs.~\ref{fig:qual}--\ref{fig:demos} probe trajectory control on
OOD letter-shaped targets and contrast unconditional vs.\
in-distribution multi-joint control; Fig.~\ref{fig:extra_demos}
shows additional patterns. Per-letter videos and multi-joint stress
demos are in the supplement. The letter trajectories test long-horizon
root-path tracking with sharp turns and self-crossing curves outside
the training prompt distribution; the same prompt threads visibly distinct paths whose
gait, foot contact, and limb posture remain text-conditioned. The
unconditional vs.\ controlled contrast shows that the K/V adapter
complements a competent frozen generator: for the same prompt, motion
semantics persist with or without trajectory control---only the
spatial path changes. The multi-joint demos stress simultaneous pelvis
and end-effector constraints where text intent and geometric targets
compete; failures concentrate at foot rows under dense per-frame
contact constraints, consistent with the foot-FID breach analysed in
the supplement.


\section{Conclusion}
\label{sec:conclusion}

We presented \textbf{KV-Control}, a parameter-efficient K/V adapter
(\paramsOursMech{} mechanism, \paramsOursTotal{} total) for trajectory
control on a co-designed PartVQ\,$+$\,T-Concat substrate. Under the
inherited refinement protocol, K/V reaches \KPSours\,cm pelvis error (M3)
and \KPSmultiBest\,cm multi-joint. At the MaskControl-matched M2 setting,
a same-backbone duplicated-branch sanity check needs $39.3\,$M
mechanism parameters to reach $1.24\,$cm, versus
$\KPSoursMaskControlProto\,$cm for our $1.5\,$M mechanism
($\sim\!26.2{\times}$ fewer).

Several directions remain open. \emph{Substrate transfer}: porting
K/V to other discrete-latent motion backbones (autoregressive,
diffusion-based, hybrid masked-AR) beyond HumanML3D's
masked-transformer family. \emph{Denser constraints}: dense per-frame
contact and multi-joint targets stress Stage-2 toward over-pinning;
tighter regularization would broaden the effective regime.
\emph{Cross-skeleton adaptation}: the PartVQ partition is
single-skeleton, and partition-aware re-attachment could lift the
adapter to other morphologies.

\bibliographystyle{ACM-Reference-Format}
\setlength{\bibsep}{0pt plus 0.2ex}
\renewcommand*{\bibfont}{\scriptsize}
\bibliography{references}

@inproceedings{momask2024,
  title={{MoMask}: Generative Masked Modeling of {3D} Human Motions},
  author={Guo, Chuan and Mu, Yuxuan and Javed, Muhammad Gohar and Wang, Sen and Cheng, Li},
  booktitle={Proceedings of the IEEE/CVF Conference on Computer Vision and Pattern Recognition (CVPR)},
  year={2024}
}

@inproceedings{mmm2024,
  title={{MMM}: Generative Masked Motion Model},
  author={Pinyoanuntapong, Ekkasit and Wang, Pu and Lee, Minwoo and Chen, Chen},
  booktitle={Proceedings of the IEEE/CVF Conference on Computer Vision and Pattern Recognition (CVPR)},
  year={2024}
}

@inproceedings{mdm2023,
  title={Human Motion Diffusion Model},
  author={Tevet, Guy and Raab, Sigal and Gordon, Brian and Shafir, Yonatan and Cohen-Or, Daniel and Bermano, Amit H.},
  booktitle={International Conference on Learning Representations (ICLR)},
  year={2023}
}

@inproceedings{mld2023,
  title={Executing your Commands via Motion Diffusion in Latent Space},
  author={Chen, Xin and Jiang, Biao and Liu, Wen and Huang, Zilong and Fu, Bin and Chen, Tao and Yu, Gang},
  booktitle={Proceedings of the IEEE/CVF Conference on Computer Vision and Pattern Recognition (CVPR)},
  year={2023}
}

@article{motiondiffuse2024,
  title={{MotionDiffuse}: Text-Driven Human Motion Generation with Diffusion Model},
  author={Zhang, Mingyuan and Cai, Zhongang and Pan, Liang and Hong, Fangzhou and Guo, Xinying and Yang, Lei and Liu, Ziwei},
  journal={IEEE Transactions on Pattern Analysis and Machine Intelligence},
  year={2024}
}

@inproceedings{t2mgpt2023,
  title={Generating Human Motion from Textual Descriptions with Discrete Representations},
  author={Zhang, Jianrong and Zhang, Yangsong and Cun, Xiaodong and Zhang, Yong and Zhao, Hongwei and Lu, Hongtao and Shen, Xi and Shan, Ying},
  booktitle={Proceedings of the IEEE/CVF Conference on Computer Vision and Pattern Recognition (CVPR)},
  year={2023}
}

@misc{acmdm2025,
  title={Absolute Coordinates Make Motion Generation Easy},
  author={Meng, Zichong and Han, Zeyu and Peng, Xiaogang and Xie, Yiming and Jiang, Huaizu},
  year={2025},
  eprint={2505.19377},
  archivePrefix={arXiv},
  primaryClass={cs.CV}
}

@inproceedings{mardm2025,
  title={Rethinking Diffusion for Text-Driven Human Motion Generation: Redundant Representations, Evaluation, and Masked Autoregression},
  author={Meng, Zichong and Xie, Yiming and Peng, Xiaogang and Han, Zeyu and Jiang, Huaizu},
  booktitle={Proceedings of the IEEE/CVF Conference on Computer Vision and Pattern Recognition (CVPR)},
  year={2025}
}

@article{motiongraphs2002,
  title={Motion Graphs},
  author={Kovar, Lucas and Gleicher, Michael and Pighin, Fr{\'e}d{\'e}ric},
  journal={ACM Transactions on Graphics},
  volume={21},
  number={3},
  pages={473--482},
  year={2002}
}

@article{pfnn2017,
  title={Phase-Functioned Neural Networks for Character Control},
  author={Holden, Daniel and Komura, Taku and Saito, Jun},
  journal={ACM Transactions on Graphics},
  volume={36},
  number={4},
  year={2017}
}

@article{lmm2020,
  title={Learned Motion Matching},
  author={Holden, Daniel and Kanoun, Oussama and Perepichka, Maksym and Popa, Tiberiu},
  journal={ACM Transactions on Graphics},
  volume={39},
  number={4},
  year={2020}
}

@inproceedings{maskcontrol2025,
  title={{MaskControl}: Spatio-Temporal Control for Masked Motion Synthesis},
  author={Pinyoanuntapong, Ekkasit and Saleem, Muhammad Usama and Karunratanakul, Korrawe and Wang, Pu and Xue, Hongfei and Chen, Chen and Guo, Chuan and Cao, Junli and Ren, Jian and Tulyakov, Sergey},
  booktitle={Proceedings of the IEEE/CVF International Conference on Computer Vision (ICCV)},
  year={2025}
}

@inproceedings{tlcontrol2024,
  title={{TLControl}: Trajectory and Language Control for Human Motion Synthesis},
  author={Wan, Weilin and Dou, Zhiyang and Komura, Taku and Wang, Wenping and Jayaraman, Dinesh and Liu, Lingjie},
  booktitle={European Conference on Computer Vision (ECCV)},
  year={2024}
}

@inproceedings{omnicontrol2024,
  title={{OmniControl}: Control Any Joint at Any Time for Human Motion Generation},
  author={Xie, Yiming and Jampani, Varun and Zhong, Lei and Sun, Deqing and Jiang, Huaizu},
  booktitle={International Conference on Learning Representations (ICLR)},
  year={2024}
}

@inproceedings{gmd2023,
  title={Guided Motion Diffusion for Controllable Human Motion Synthesis},
  author={Karunratanakul, Korrawe and Preechakul, Konpat and Suwajanakorn, Supasorn and Tang, Siyu},
  booktitle={Proceedings of the IEEE/CVF International Conference on Computer Vision (ICCV)},
  year={2023}
}

@inproceedings{condmdi2024,
  title={Flexible Motion In-betweening with Diffusion Models},
  author={Cohan, Setareh and Tevet, Guy and Reda, Daniele and Peng, Xue Bin and van de Panne, Michiel},
  booktitle={ACM SIGGRAPH 2024 Conference Papers},
  year={2024}
}

@inproceedings{motionlcm2024,
  title={{MotionLCM}: Real-time Controllable Motion Generation via Latent Consistency Model},
  author={Dai, Wenxun and Chen, Ling-Hao and Wang, Jingbo and Liu, Jinpeng and Dai, Bo and Tang, Yansong},
  booktitle={European Conference on Computer Vision (ECCV)},
  year={2024}
}

@inproceedings{vqvae2017,
  title={Neural Discrete Representation Learning},
  author={van den Oord, Aaron and Vinyals, Oriol and Kavukcuoglu, Koray},
  booktitle={Advances in Neural Information Processing Systems (NeurIPS)},
  year={2017}
}

@inproceedings{clip2021,
  title={Learning Transferable Visual Models From Natural Language Supervision},
  author={Radford, Alec and Kim, Jong Wook and Hallacy, Chris and Ramesh, Aditya and Goh, Gabriel and Agarwal, Sandhini and Sastry, Girish and Askell, Amanda and Mishkin, Pamela and Clark, Jack and Krueger, Gretchen and Sutskever, Ilya},
  booktitle={International Conference on Machine Learning (ICML)},
  year={2021}
}

@inproceedings{humanml3d2022,
  title={Generating Diverse and Natural 3D Human Motions from Text},
  author={Guo, Chuan and Zou, Shihao and Zuo, Xinxin and Wang, Sen and Ji, Wei and Li, Xingyu and Cheng, Li},
  booktitle={Proceedings of the IEEE/CVF Conference on Computer Vision and Pattern Recognition (CVPR)},
  year={2022}
}

@inproceedings{controlnet2023,
  title={Adding Conditional Control to Text-to-Image Diffusion Models},
  author={Zhang, Lvmin and Rao, Anyi and Agrawala, Maneesh},
  booktitle={Proceedings of the IEEE/CVF International Conference on Computer Vision (ICCV)},
  year={2023}
}

@inproceedings{lora2022,
  title={{LoRA}: Low-Rank Adaptation of Large Language Models},
  author={Hu, Edward J. and Shen, Yelong and Wallis, Phillip and Allen-Zhu, Zeyuan and Li, Yuanzhi and Wang, Shean and Wang, Lu and Chen, Weizhu},
  booktitle={International Conference on Learning Representations (ICLR)},
  year={2022}
}

@inproceedings{li2021prefix,
  title={Prefix-Tuning: Optimizing Continuous Prompts for Generation},
  author={Li, Xiang Lisa and Liang, Percy},
  booktitle={Proceedings of the 59th Annual Meeting of the Association for Computational Linguistics and the 11th International Joint Conference on Natural Language Processing},
  year={2021},
  pages={4582--4597}
}

@misc{ye2023ipadapter,
  title={IP-Adapter: Text Compatible Image Prompt Adapter for Text-to-Image Diffusion Models},
  author={Ye, Hu and Zhang, Jun and Liu, Sibo and Han, Xiao and Yang, Wei},
  year={2023},
  eprint={2308.06721},
  archivePrefix={arXiv},
  primaryClass={cs.CV}
}

@misc{ho2022cfg,
  title={Classifier-Free Diffusion Guidance},
  author={Ho, Jonathan and Salimans, Tim},
  year={2022},
  eprint={2207.12598},
  archivePrefix={arXiv},
  primaryClass={cs.LG}
}

@inproceedings{motiongpt2023,
  title={{MotionGPT}: Human Motion as a Foreign Language},
  author={Jiang, Biao and Chen, Xin and Liu, Wen and Yu, Jingyi and Yu, Gang and Chen, Tao},
  booktitle={Advances in Neural Information Processing Systems (NeurIPS)},
  year={2023}
}

@inproceedings{remodiffuse2023,
  title={{ReMoDiffuse}: Retrieval-Augmented Motion Diffusion Model},
  author={Zhang, Mingyuan and Guo, Xinying and Pan, Liang and Cai, Zhongang and Hong, Fangzhou and Li, Huirong and Yang, Lei and Liu, Ziwei},
  booktitle={International Conference on Computer Vision (ICCV)},
  year={2023}
}

@inproceedings{bamm2024,
  title={{BAMM}: Bidirectional Autoregressive Motion Model},
  author={Pinyoanuntapong, Ekkasit and Saleem, Muhammad Usama and Wang, Pu and Lee, Minwoo and Das, Srijan and Chen, Chen},
  booktitle={European Conference on Computer Vision (ECCV)},
  year={2024}
}

@inproceedings{motionmamba2024,
  title={Motion Mamba: Efficient and Long Sequence Motion Generation},
  author={Zhang, Zeyu and Liu, Akide and Reid, Ian and Hartley, Richard and Zhuang, Bohan and Tang, Hao},
  booktitle={European Conference on Computer Vision (ECCV)},
  year={2024}
}

@inproceedings{finemogen2023,
  title={{FineMoGen}: Fine-Grained Spatio-Temporal Motion Generation and Editing},
  author={Zhang, Mingyuan and Li, Huirong and Cai, Zhongang and Ren, Jiawei and Yang, Lei and Liu, Ziwei},
  booktitle={Advances in Neural Information Processing Systems (NeurIPS)},
  year={2023}
}

@inproceedings{priormdm2024,
  title={Human Motion Diffusion as a Generative Prior},
  author={Shafir, Yonatan and Tevet, Guy and Kapon, Roy and Bermano, Amit H.},
  booktitle={International Conference on Learning Representations (ICLR)},
  year={2024}
}

@inproceedings{dno2024,
  title={Optimizing Diffusion Noise Can Serve As Universal Motion Priors},
  author={Karunratanakul, Korrawe and Preechakul, Konpat and Aksan, Emre and Beeler, Thabo and Suwajanakorn, Supasorn and Tang, Siyu},
  booktitle={IEEE/CVF Conference on Computer Vision and Pattern Recognition (CVPR)},
  year={2024}
}

@inproceedings{intercontrol2024,
  title={{InterControl}: Zero-shot Human Interaction Generation by Controlling Every Joint},
  author={Wang, Zhenzhi and Wang, Jingbo and Li, Yixuan and Lin, Dahua and Dai, Bo},
  booktitle={Advances in Neural Information Processing Systems (NeurIPS)},
  year={2024}
}

@inproceedings{como2024,
  title={{CoMo}: Controllable Motion Generation through Language Guided Pose Code Editing},
  author={Huang, Yiming and Wan, Weilin and Yang, Yue and Callison-Burch, Chris and Yatskar, Mark and Liu, Lingjie},
  booktitle={European Conference on Computer Vision (ECCV)},
  year={2024}
}

@inproceedings{m2d2m2024,
  title={{M2D2M}: Multi-Motion Generation from Text with Discrete Diffusion Models},
  author={Chi, Seunggeun and Chi, Hyung-gun and Ma, Hengbo and Agarwal, Nakul and Siddiqui, Faizan and Ramani, Karthik and Lee, Kwonjoon},
  booktitle={European Conference on Computer Vision (ECCV)},
  year={2024}
}

@inproceedings{lightt2m2025,
  title={{Light-T2M}: A Lightweight and Fast Model for Text-to-Motion Generation},
  author={Zeng, Ling-An and Huang, Guohong and Wu, Gaojie and Zheng, Wei-Shi},
  booktitle={AAAI Conference on Artificial Intelligence},
  year={2025}
}

@inproceedings{parco2024,
  title={{ParCo}: Part-Coordinating Text-to-Motion Synthesis},
  author={Zou, Qiran and Yuan, Shangyuan and Du, Shian and Wang, Yu and Liu, Chang and Xu, Yi and Chen, Jie and Ji, Xiangyang},
  booktitle={European Conference on Computer Vision (ECCV)},
  pages={126--143},
  publisher={Springer},
  address={Milan, Italy},
  year={2024}
}

@inproceedings{maskgit2022,
  title={{MaskGIT}: Masked Generative Image Transformer},
  author={Chang, Huiwen and Zhang, Han and Jiang, Lu and Liu, Ce and Freeman, William T.},
  booktitle={IEEE/CVF Conference on Computer Vision and Pattern Recognition (CVPR)},
  year={2022}
}

@inproceedings{adaptertuning2019,
  title={Parameter-Efficient Transfer Learning for {NLP}},
  author={Houlsby, Neil and Giurgiu, Andrei and Jastrzebski, Stanislaw and Morrone, Bruna and de Laroussilhe, Quentin and Gesmundo, Andrea and Attariyan, Mona and Gelly, Sylvain},
  booktitle={International Conference on Machine Learning (ICML)},
  year={2019}
}


\begin{figure*}[!p]
\centering
\includegraphics[width=0.86\textwidth]{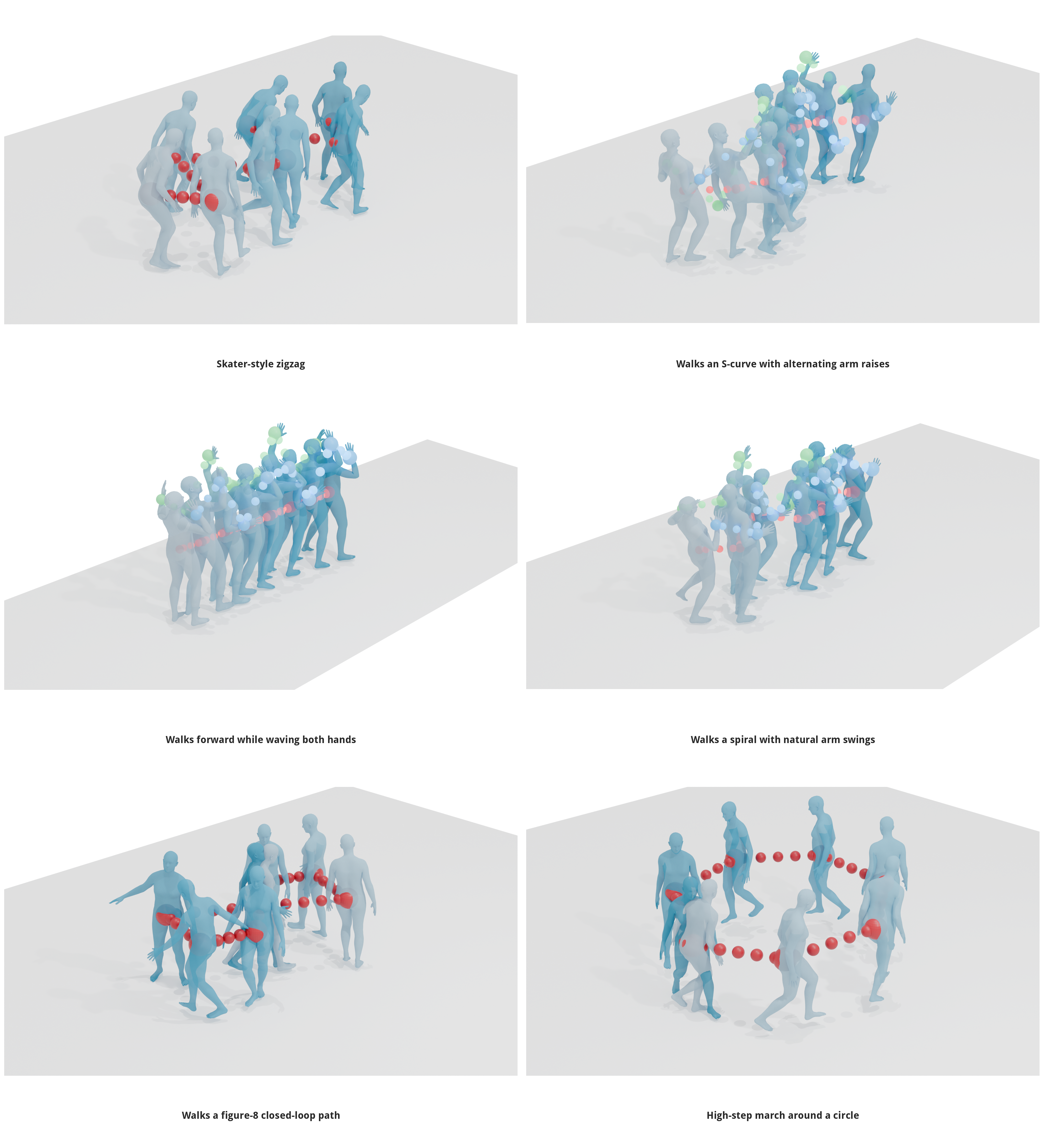}
\caption{\textbf{Extended qualitative behaviour of the same trained
K/V-Control adapter.} Six probes beyond Figs.~\ref{fig:qual}--\ref{fig:demos},
all using the same multi-joint K/V-Control checkpoint. \emph{Row 1:} skater-style
zigzag (single-joint) and an S-curve walk with alternating arm raises (multi-joint).
\emph{Row 2:} forward locomotion with both hands waving and a spiral walk with
natural arm swings (both multi-joint). \emph{Row 3:} a figure-eight closed-loop
path and a high-step march around a circle (both single-joint). Colored 3D markers
are the control targets (red\,$=$\,pelvis, blue\,$=$\,l\_wrist, green\,$=$\,r\_wrist);
per-demo videos in the supplementary.}
\label{fig:extra_demos}
\end{figure*}

\end{document}